# 基于高阶一致性学习的聚类集成算法

甘舰文[1],陈　艳[2],周　芃[3],杜　亮[1,4*]
(1. 山西大学 计算机与信息技术学院,太原 030006; 2. 四川大学 计算机学院,成都 610065;
3. 安徽大学 计算机科学与技术学院,合肥 230601; 4. 山西大学 大数据科学与产业研究院,太原 030006)
(*通信作者电子邮箱duliang@sxu.edu.cn)

**摘　要**:现有的大部分关于聚类集成的研究主要关注有效的集成算法的设计。为解决由于基聚类器的质量高低不一、低质量的基聚类器对聚类集成性能产生影响的问题,从数据发掘的角度出发,以基聚类器为基础挖掘数据的内在联系,提出一种高阶信息融合算法——基于高阶一致性学习的聚类集成(HCLCE)算法,从不同的维度表示数据之间的联系。首先,将每种高阶信息融合成一个新的结构化的一致性矩阵;然后,再对得到的多个一致性矩阵进行融合;最后,将多种信息融合为一个一致性的结果。实验结果表明,与次优的LWEA(Locally Weighted Evidence Accumulation)算法相比,HCLCE算法的聚类准确率平均提升了7.22%,归一化互信息(NMI)平均提升了9.19%。可见,HCLCE能得到比聚类集成算法和单独使用一种信息更好的聚类结果。

## Clustering ensemble algorithm with high-order consistency learning

GAN Jianwen[1], CHEN Yan[2], ZHOU Peng[3], DU Liang[1,4*]
(1. *School of Computer and Information Technology*, *Shanxi University*, *Taiyuan Shanxi* 030006, *China*;
2. *College of Computer Science*, *Sichuan University*, *Chengdu Sichuan* 610065, *China*;
3. *School of Computer Science and Technology*, *Anhui University*, *Hefei Anhui* 230601, *China*;
4. *Institute of Big Data Science and Industry*, *Shanxi University*, *Taiyuan Shanxi* 030006, *China*)

**Abstract:** Most of the research on clustering ensemble focuses on designing practical consistency learning algorithms. To solve the problems that the quality of base clusters varies and the low-quality base clusters have an impact on the performance of the clustering ensemble, from the perspective of data mining, the intrinsic connections of data were mined based on the base clusters, and a high-order information fusion algorithm was proposed to represent the connections between data from different dimensions, namely Clustering Ensemble with High-order Consensus learning (HCLCE). Firstly, each high-order information was fused into a new structured consistency matrix. Then, the obtained multiple consistency matrices were fused together. Finally, multiple information was fused into a consistent result. Experimental results show that LCLCE algorithm has the clustering accuracy improved by an average of 7.22%, and the Normalized Mutual Information (NMI) improved by an average of 9.19% compared with the suboptimal Locally Weighted Evidence Accumulation (LWEA) algorithm. It can be seen that the proposed algorithm can obtain better clustering results compared with clustering ensemble algorithms and using one information alone.

## 0　引言

聚类是一种重要的无监督分类技术,在统计、模式识别、机器学习、数据挖掘等不同领域都得到了广泛的研究。根据几种聚类准则和不同的相似性度量方法,可以揭示一个数据集的底层结构。在无监督学习中,由于训练数据集没有标签,聚类算法很难验证聚类结果的有效性,给设计聚类算法带来很大的挑战。每种聚类算法都有自己的优缺点,传统的聚类算法同时还面临其他问题[1],例如同一种聚类算法,由于目标函数的不同,在相同的数据集上也会得到不同的聚类结果。K-均值(K-Means)算法高度依赖初始化和数据分布。为了提高单个聚类算法结果的鲁棒性、一致性、新颖性和稳定性,聚类集成(聚类融合或共识聚类)利用多个聚类结果的共识并将它们组合成最优解。聚类集成提供了一种框架,可以将多个基聚类器的结果组合在一起,生成一致聚类[1]。

现有的聚类算法可以被分为三类:

1)基于相似性矩阵的算法。将基聚类结果转化为相似性矩阵,通过不同的聚类集成方法生成一致性矩阵。

**收稿日期**:2022-09-12;**修回日期**:2022-10-28;**录用日期**:2022-11-07。　**基金项目**:国家自然科学基金资助项目(61976129)。

**作者简介**:甘舰文(1996—),男,河南商丘人,硕士研究生,主要研究方向:聚类集成、数据挖掘;　陈艳(1994—),女,山西太原人,博士研究生,主要研究方向:多核聚类、深度聚类;　周芃(1989—),男,安徽合肥人,副教授,博士,主要研究方向:聚类集成、数据挖掘;　杜亮(1985—),男,山西太原人,副教授,博士,主要研究方向:机器学习、大数据分析。

2)图方法。将输入的基聚类结果转化为无向图,通过图划分得到最终的聚类结果。

3)基于重标记的方法。将基聚类结果转化为新的标签向量,然后通过标签对齐找到集合聚类。

基于相似性矩阵的方法和图方法近年来有着广泛的应用。相似性矩阵反映样本对之间的关系,在聚类集成算法中使用广泛。不同的相似度度量方式会得到不同的结果。但这两种方法的输入数据易受离群点影响而破坏簇的边界,从而影响到最终聚类结果[2]。本文通过基聚类器生成相似性矩阵,从不同的角度度量样本对之间的相似性。

## 1 相关工作

聚类集成融合多种输入结果试图得到一个更好的结果,至今已经发展出一大批聚类集成方法。在聚类集成方法发展早期,一些以信息论为基础的方法被提出,如Strehl等[3]以信息共享为基础,将聚类集成问题转化为组合优化问题。近年来,更多的方法被应用到聚类集成中,如利用对齐的方法结合多个K-Means的聚类结果[4]。一些工作利用非负矩阵分解将关联矩阵分解为两个指示矩阵[5]。除了以K-Means作为基聚类输入,谱聚类也有聚类集成的工作。一些方法引入了概率理论将图模型转化为聚类集合,如Wang等[6]使用了贝叶斯模型聚类集成学习了一个带有因子图的共识聚类结果。

由于聚类的多样性和质量在集成学习中至关重要,许多方法都充分利用多样性和质量来组合基聚类。如Abbasi等[7]提出了一种新的稳定性测度——归一化互信息(Normalized Mutual Information, NMI),并将它用于集合基聚类;Bagherinia等[8]考虑基聚类结果的多样性和质量提出了一种模糊聚类集合。除了使用所有基聚类器的结果作为输入进行聚类集成,还有一些工作试图选择一些具有高质量信息且无冗余的基聚类结果进行集成。Azimi等[9]提出了一种自适应聚类集合选择方法来选择基聚类结果;Hong等[10]采用重采样方法选择基聚类;Parvin等[11]提出了一种加权局部自适应聚类集合选择算法;Yu等[12]将聚类选择转化为特征选择,设计了一种混合策略来选择基聚类结果;Zhao等[13]提出了用于聚类集合选择的内部有效性指标;Shi等[14]将迁移学习扩展到聚类集成,提出了迁移聚类集成选择方法。

根据算法思想和原理这些聚类集成方法可归类为:基于关系矩阵的方法、直接融合法和基于图的方法。Li等[15]提出了规范化边的概念用来度量样本的相似度,用层次聚类来融合最终的结果;Huang等[16]使用概率轨迹的概念重新构造样本相似度。直接融合法首先匹配基聚类器中的类簇,然后通过投票机制融合结果。图方法在基聚类器上构建图表示,利用图分割技术发现群组结构。常用的图划分技术包括归一化切割(Normalized CUT, N-CUT)[17]和层次化的分割算法METIS[18]。聚类方法的设计和输入的基聚类结果都会显著影响聚类集成的性能。基聚类结果应该尽可能地体现差异性,而不是追求数量。获得差异性的基聚类结果主要有以下几种方式:1)使用不同的聚类方法对同一数据集进行聚类;2)使用不同的初始化值和有差异性的参数值;3)对进行聚类的数据集使用不同的办法抽样,获得有区别的数据片段。

本文方法基于相似性矩阵,一方面利用高阶信息有效地发掘数据样本之间的联系,另一方面不同角度的信息使得参与融合的基聚类信息具有较大的差异性。同时,利用多种信息源也会带来处理高阶数据耗时长、计算量大的问题。针对以上问题本文提出一种新的高阶信息融合算法——基于高阶一致性学习的聚类集成(Clustering Ensemble with High-order Consensus Learning, HCLCE)算法。首先将每种高阶信息融合成一个新的结构化的一致性矩阵;然后再对得到的多个一致性矩阵进行融合。算法通过双随机约束,使得一致性矩阵行列求和的值都为1,因此样本对之间的相似度,同时也表示了该样本与其他样本属于同一个类的概率。

## 2 基于高阶一致性学习的聚类集成

本章首先介绍高阶信息的表示方法,然后描述HCLCE算法的具体细节,最后对目标函数进行求解优化。

$\boldsymbol{X}=\{\boldsymbol{X}_1,\boldsymbol{X}_2,\cdots,\boldsymbol{X}_i,\cdots,\boldsymbol{X}_m\}$为$d$维空间中未标记的$n$个样本,通过K-Means算法进行$m$次聚类,生成基聚类结果$\boldsymbol{H}=\{\boldsymbol{H}_1,\boldsymbol{H}_2,\cdots,\boldsymbol{H}_i,\cdots,\boldsymbol{H}_m\}$,其中:$\boldsymbol{H}_i$表示第$i$次聚类的结果;$c$表示簇的个数,假设所有基聚类器结果簇的个数一样。基于$\boldsymbol{H}_i$,相似性矩阵$\boldsymbol{S}^i$可以表示为:$\boldsymbol{S}^i=\boldsymbol{H}_i\boldsymbol{H}_i^{\mathrm{T}}$,同时定义$\mathbf{1}$表示大小为$n\times 1$的列向量。

### 2.1 高阶矩阵信息定义

单个基聚类器相似性矩阵是一次聚类的结果,为了挖掘样本之间进一步的联系,利用多次聚类的结果,获取更具有代表性和差异性的高阶信息。本文通过以下几种方式,从不同的角度获得聚类信息增益。

#### 2.1.1 一阶一致性

单次聚类结果的相似性矩阵结果$\boldsymbol{S}^i$之间差异性较小。以$\boldsymbol{A}_i^1=\boldsymbol{S}^i$表示把单个相似性矩阵作为第一种输入信息。

加权结构化的过程可以分为两步,由于一阶信息是由$m$个聚类共识结果组成,每个聚类结果之间具有一定差异性,因此第一步对集合$\boldsymbol{A}^1$中的每个相似性矩阵赋予权重,融合成一个相似性矩阵$\hat{\boldsymbol{S}}^1$,表示为:

$$\max_{\boldsymbol{w},\hat{\boldsymbol{S}}^1}\sum_{i=1}^{k}w_i\mathrm{Tr}\left(\boldsymbol{A}_i^1\hat{\boldsymbol{S}}^1\right) \tag{1}$$

$$\text{s.t.}\quad \sum_{i=1}^{k}w_i^2=1,\ w_i\geqslant 0,\ \hat{\boldsymbol{S}}^1\geqslant 0,\ \sum_{j=1}^{n}\left(\hat{S}_{ij}^1\right)^2=1;\ \forall i$$

其中:$k$是集合$\boldsymbol{A}^1$中元素的数量,在一阶情况下$k$的大小等于输入的相似性矩阵的个数;$\hat{\boldsymbol{S}}^1$是$k$个相似性矩阵加权融合的结果,$\hat{S}_{ij}^1$为矩阵中的元素;$w_i$是权重向量$\boldsymbol{w}$的第$i$个元素。通过对$\hat{\boldsymbol{S}}^1$结构化使簇的结构更清楚,同时满足相似性矩阵性质的约束。

对$\hat{\boldsymbol{S}}^1$结构化的过程[19]为:

$$\min_{\boldsymbol{M}^1,\boldsymbol{F}}\left\|\boldsymbol{M}^1-\hat{\boldsymbol{S}}^1\right\|_F^2+2\lambda\mathrm{Tr}\left(\boldsymbol{F}^{\mathrm{T}}\boldsymbol{L}\boldsymbol{F}\right) \tag{2}$$

$$\text{s.t.}\quad \boldsymbol{M}^1\geqslant 0,\ \boldsymbol{M}^1=(\boldsymbol{M}^1)^{\mathrm{T}},\ \boldsymbol{M}^1\mathbf{1}=\mathbf{1},\ \boldsymbol{F}\in\mathbf{R}^{n\times c},\ \boldsymbol{F}^{\mathrm{T}}\boldsymbol{F}=\boldsymbol{I}$$

其中:$\boldsymbol{L}$是拉普拉斯矩阵;$\lambda$是自适应参数。

求得的$\boldsymbol{M}^1\mathbf{1}$对称且满足双随机约束,是一阶信息加权结构化后的一致性矩阵。

#### 2.1.2 二阶簇级一致性

二阶簇级一致性表示两个基聚类器对同一个簇的一致性进行投票。得分越大,说明不同基聚类器之间同一个样本所在的两个簇之间交集越大,越具有相似性。簇的一致性的投票计算过程如图1所示,可以表示为:$\boldsymbol{A}_{ij}^2=\boldsymbol{S}^i\boldsymbol{S}^j$。

二阶簇级一致性是基聚类器两两运算,基聚类器之间不进行运算,所以$m$个输入会产生$m^2-m$个结果,而相似性矩

阵本身对称，因此只需要计算$(m^2-m)/2$次，对$A^2$加权得：

$$\max_{w,\hat{S}^2}\sum_{i=1}^{m^2-m} w_i\mathrm{Tr}\left(A_i^2\hat{S}^2\right) \tag{3}$$

$$\text{s. t.}\ \sum_{i=1}^{m^2-m} w_i^2=1,\ \hat{S}^2\geqslant 0,\ \sum_{j=1}^{n}\left(\hat{S}_{ij}^2\right)^2=1;\ \forall i$$

对$\hat{S}^2$结构化的过程为：

$$\min_{M^2,F}\left\|M^2-\hat{S}^2\right\|_F^2+2\lambda\mathrm{Tr}\left(F^{\mathrm{T}}LF\right) \tag{4}$$

$$\text{s. t.}\ \ M^2\geqslant 0,\ M^2=(M^2)^{\mathrm{T}},\ M^2\mathbf{1}=\mathbf{1},\ F\in\mathrm{R}^{n\times c},\ F^{\mathrm{T}}F=I$$

求得$M^2$对称且满足双随机约束，表示二阶的簇级信息加权结构化后的一致性矩阵。

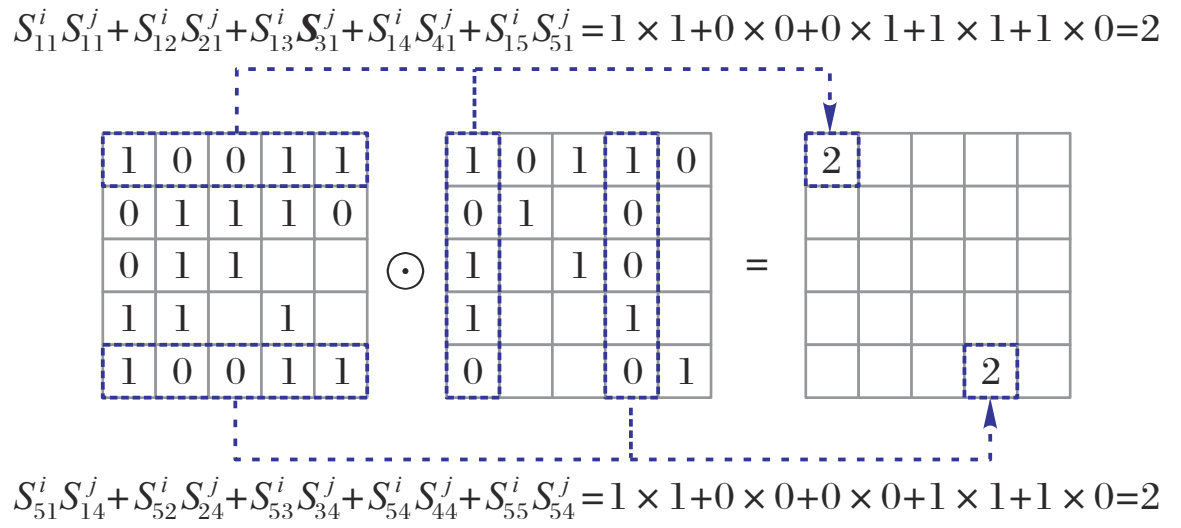

图1　相似性矩阵$S^i$和$S^j$簇间交集大小的计算

Fig. 1　Calculation of intersection size between clusters of similarity matrix $S^i$ and $S^j$

### 2.1.3　二阶样本对一致性

相似性矩阵的每一个元素的值代表着不同样本对两两之间一致性的大小。通过相似性矩阵两两之间进行点乘运算，只有在两种样本对取值都为1的情况下，样本对是否属于一个类才能达成一致，否则认为不属于同一类，这说明点乘运算只会保留达成一致的样本对，不一致的样本将会舍去，计算过程如图2所示，相似性矩阵对同一个样本对相似性进行乘法运算，显然只有达成一致的样本对相似性为1，否则为0。所以这种情况下的高阶信息同时增强了样本对之间的确定性和不确定性，表示为：$A_{ij}^3=S^i\odot S^j$。对于$m$个相似性矩阵哈达玛积也会产生$(m^2-m)/2$个结果。

对$A^3$加权得：

$$\max_{w,\hat{S}^3}\sum_{i=1}^{m^2-m} w_i\mathrm{Tr}\left(A_i^3\hat{S}^3\right) \tag{5}$$

$$\text{s. t.}\ \sum_{i=1}^{m^2-m} w_i^2=1,\ w_i\geqslant 0,\ \hat{S}^3\geqslant 0,\ \sum_{j=1}^{n}\left(\hat{S}_{ij}^3\right)^2=1;\ \forall i$$

对$\hat{S}^3$结构化的过程为：

$$\min_{M^3,F}\left\|M^3-\hat{S}^3\right\|_F^2+2\lambda\mathrm{Tr}\left(F^{\mathrm{T}}LF\right) \tag{6}$$

$$\text{s. t.}\ \ M^3\geqslant 0,\ M^3=(M^3)^{\mathrm{T}},\ M^3\mathbf{1}=\mathbf{1},\ F\in\mathrm{R}^{n\times c},\ F^{\mathrm{T}}F=I$$

求得$M^3$对称且满足双随机约束，是二阶样本对之间信息加权结构化的一致性矩阵。

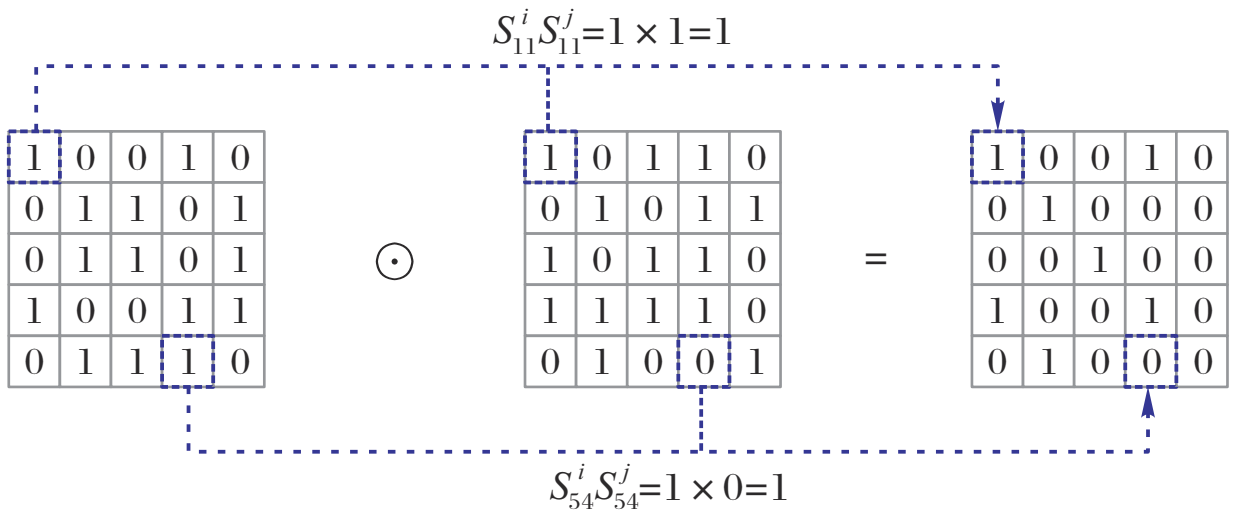

图2　相似性矩阵$S^i$和$S^j$样本对的一致性计算

Fig. 2　Consistency calculation of similarity matrix $S^i$ and $S^j$ sample pair

### 2.1.4　$m$阶样本对一致性

在此基础上，可以提出一种更加严格的样本对一致性信息挖掘方式，表示为：$A^4=\prod_{i=1}^{m}S^1\odot S^2\odot\cdots\odot S^m$，这种运算表示对所有单次聚类结果进行连乘点积运算，只有所有结果达成一致的样本对才会被保留，任何一个相似性矩阵的不一致结果，都会使该样本对结果为0，计算过程如图3。可以看到，对不同相似性矩阵中的样本对相似度相乘，只有所有相似性矩阵在该样本对上的值为1时，得到的最终矩阵才会保留该样本对相似度为1。因此，保留下的样本对具有最大的确定性，同时该矩阵也最稀疏。$A^4$最后结果只有一个矩阵，因此不需要赋予权重。定义$S^4=A^4$，对$\hat{S}^4$结构化的过程为：

$$\min_{M^4,F}\left\|M^4-\hat{S}^4\right\|_F^2+2*\lambda\mathrm{Tr}(F^{\mathrm{T}}LF) \tag{7}$$

$$\text{s. t.}\ \ M^4\geqslant 0,\ M^4=(M^4)^{\mathrm{T}},\ M^4\mathbf{1}=\mathbf{1},\ F\in\mathrm{R}^{n\times c},\ F^{\mathrm{T}}F=I$$

求得$M^4$对称且满足双随机约束，是$m$阶样本对级别的信息加权结构化后的一致性矩阵。

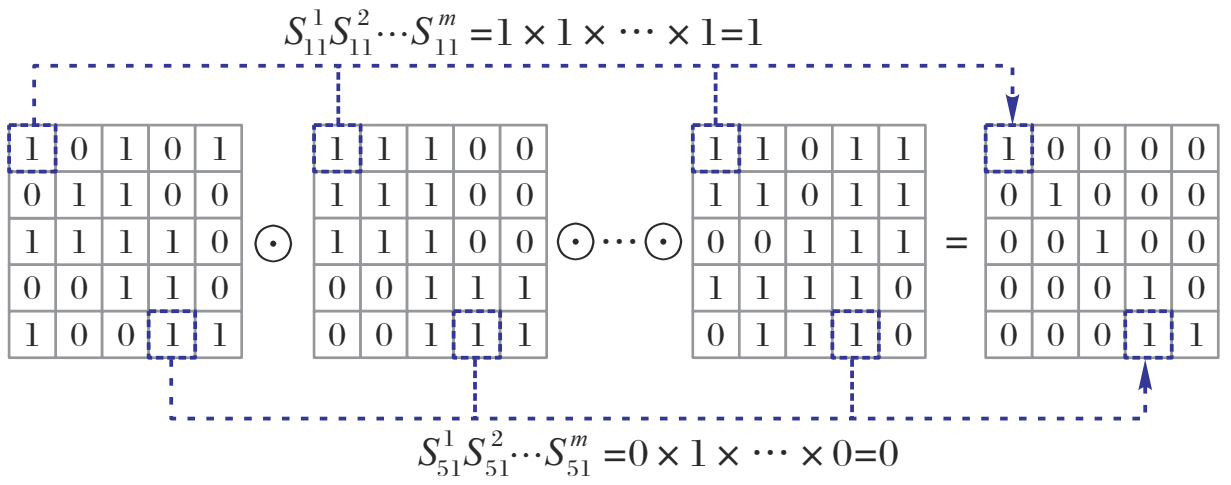

图3　所有相似性矩阵样本对一致性计算

Fig. 3　Consistency calculation for all similarity matrix sample pairs

### 2.1.5　高阶信息融合

聚类集成将多个共识结果组合为一个最优解，由于对高阶信息的发掘，特别是相似性矩阵两两之间的关联，使得需要融合的共识结果迅速增多，一次性融合这些信息需要耗费巨大的时间和计算量，为此本文提出一种分阶段融合的框架。对每种高阶信息进行计算，先融合成一种加权结构化后的高阶信息，将它作为输入，最终融合为一个一致性矩阵。用$M$表示满足约束条件，是最终学习的一致性矩阵。整体算法流程如图4所示。

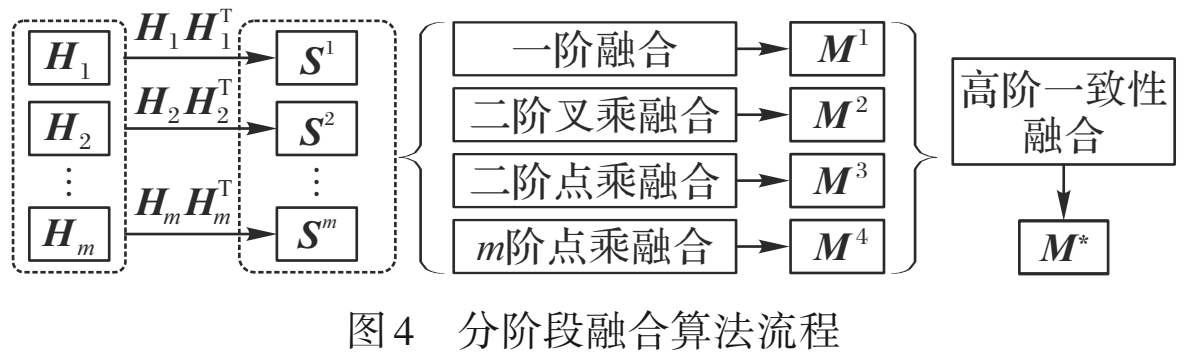

图4　分阶段融合算法流程

Fig. 4　Flowchart of phased fusion algorithm

由于每种高阶信息携带的信息和侧重点不同，为了放大信息间差异性的作用，仍需要对每种信息赋予权重，如式(8)所示：

$$\min_{M,F}\left\|M-\sum_{i=1}^{d}w_iM^i\right\|_F^2+2\lambda\mathrm{Tr}(F^{\mathrm{T}}LF) \tag{8}$$

$$\text{s. t.}\ \ M\geqslant 0,\ M=M^{\mathrm{T}},\ M\mathbf{1}=\mathbf{1},\ \sum_{i=1}^{d}w_i=1,\ F\in\mathrm{R}^{n\times c},\ F^{\mathrm{T}}F=I$$

其中：$L$是拉普拉斯矩阵，$L_M=D_M-M$，$D_M$为矩阵$M$的度矩阵，$D_M\in\mathrm{R}^{n\times n}$定义为一个对角矩阵，第$i$个元素为$\sum_j M_{ij}$，通过增加秩约束，使得$\mathrm{rank}(L_M)=n-c$，学得的一致性矩阵有

$c$个连通片，从而获得更加清晰的簇结构[19]；$d$是需要融合信息的个数；$\lambda$是自适应参数，随着$\mathrm{rank}(\boldsymbol{L}_{\boldsymbol{M}})$的大小自动调整；$\boldsymbol{M}^i$是第$i$种加权结构化后的高阶信息输入。下面介绍如何求解所提出的目标函数。

## 2.2 模型优化

本节主要介绍优化问题的求解方法和算法流程。

### 2.2.1 加权优化

优化问题式(1)、(3)、(5)为同一种问题，区别在于权重个数不同。以式(1)为例，迭代更新$\boldsymbol{w}$和$\boldsymbol{A}^1$。

1)固定$\boldsymbol{w}$，求$\hat{\boldsymbol{S}}^1$。

固定$\boldsymbol{w}$，式(1)可化简为：

$$\max_{\hat{S}^1} \mathrm{Tr}\left(\boldsymbol{D}\hat{\boldsymbol{S}}^1\right) \tag{9}$$

$$\mathrm{s.t.}\quad \hat{\boldsymbol{S}}^1 \geqslant 0,\ \sum_{j=1}^{n}\left(S_{ij}^1\right)^2 = 1, \forall i$$

其中，$\boldsymbol{D} = \sum_i^k w_i \boldsymbol{A}_i^1$，式(9)约束条件为$\hat{\boldsymbol{S}}^1$行平方和为1，可直接通过归一化求解。易得：

$$\hat{S}_{ij}^1 = D_{ij} \Big/ \sqrt{\sum_{i'=1}^{n}\left(D_{i'j}\right)^2} \tag{10}$$

其中，$D_{ij}$为矩阵$\boldsymbol{D}$中第$i$行$j$列的元素。

2)固定$\hat{\boldsymbol{S}}^1$，求$\boldsymbol{w}$。

此时式(1)可化简为：

$$\max_{\boldsymbol{w}} \boldsymbol{w}^{\mathrm{T}}\boldsymbol{b} \tag{11}$$

$$\mathrm{s.t.}\quad \sum_{i=1}^{k} w_i^2 = 1, w_i \geqslant 0$$

其中：$\boldsymbol{b} \in \mathbf{R}^{n\times 1}$，$b_i = \mathrm{Tr}\left(\boldsymbol{A}^i\hat{\boldsymbol{S}}^1\right)$。易得：

$$w_i = b_i \Big/ \sqrt{\sum_{i'=1}^{k}(b_{i'})^2} \tag{12}$$

### 2.2.2 结构化算法

优化问题式(2)、(4)、(6)、(7)为同一类问题。以式(2)为例：

1)固定$\boldsymbol{F}$，更新$\boldsymbol{M}^1$。

$$\min_{\boldsymbol{M}^1} \|\boldsymbol{M}^1 - \boldsymbol{B}\|_{\mathrm{F}}^2 \tag{13}$$

$$\mathrm{s.t.}\quad \boldsymbol{M}^1 \geqslant 0, \boldsymbol{M}^1 = (\boldsymbol{M}^1)^{\mathrm{T}}, \boldsymbol{M}^1\mathbf{1} = \mathbf{1}$$

其中，$\boldsymbol{B} \in \mathbf{R}^{n\times n}$，$\boldsymbol{B} = \boldsymbol{M}^1 - \lambda\boldsymbol{D}$，$\boldsymbol{D}$是$\boldsymbol{F}$的欧氏距离矩阵。

式(2)可以拆成两个子问题。

子问题1：

$$\min_{\boldsymbol{M}^1} \|\boldsymbol{M}^1 - \boldsymbol{B}\|_{\mathrm{F}}^2 \tag{14}$$

$$\mathrm{s.t.}\quad \boldsymbol{M}^1 = (\boldsymbol{M}^1)^{\mathrm{T}}, \boldsymbol{M}^1\mathbf{1} = \mathbf{1}$$

子问题2：

$$\min_{\boldsymbol{M}^1} \|\boldsymbol{M}^1 - \boldsymbol{B}\|_{\mathrm{F}}^2 \tag{15}$$

$$\mathrm{s.t.}\quad \boldsymbol{M}^1 \geqslant 0$$

根据Von Neumann交替投影定理[20]。本文使用的这种相互投影策略将收敛于由问题(14)和(15)形成的两个子空间的交叉。对式(14)[21]求解可得：

$$\boldsymbol{M}^1 = \boldsymbol{K} + \frac{n + \mathbf{1}^{\mathrm{T}}\boldsymbol{K}\mathbf{1}}{n^2}\mathbf{1}\cdot\mathbf{1}^{\mathrm{T}} - \frac{1}{n}\boldsymbol{K}\mathbf{1}\cdot\mathbf{1}^{\mathrm{T}} - \frac{1}{n}\mathbf{1}\cdot\mathbf{1}^{\mathrm{T}}\boldsymbol{K} \tag{16}$$

其中，$\boldsymbol{K} = \left(\boldsymbol{B} + \boldsymbol{B}^{\mathrm{T}}\right)/2$。

将$\boldsymbol{M}^1$的值作为输入代入式(15)中赋予$\boldsymbol{B}$求解，易得：

$$\boldsymbol{M}^1 = \max(0, \boldsymbol{B}) \tag{17}$$

将得到$\boldsymbol{M}^1$作为$\boldsymbol{B}$代入式(14)，如此迭代直至$\boldsymbol{M}^1$收敛。

2)固定$\boldsymbol{M}^1$，更新$\boldsymbol{F}$。

优化问题(2)可化简为：

$$\min_{\boldsymbol{F}} \mathrm{Tr}(\boldsymbol{F}^{\mathrm{T}}\boldsymbol{L}\boldsymbol{F}) \tag{18}$$

$$\mathrm{s.t.}\quad \boldsymbol{F} \in \mathbf{R}^{n\times c}, \boldsymbol{F}^{\mathrm{T}}\boldsymbol{F} = \boldsymbol{I}$$

根据Ky Fan's theorem理论[22]，$\boldsymbol{F}$为$\boldsymbol{L}$前$c$个最小的特征向量。

### 2.2.3 融合算法

求解式(8)，可以迭代地更新$\boldsymbol{M}$、$\boldsymbol{W}$、$\boldsymbol{F}$，详细过程如下：

1)固定$\boldsymbol{M}$，更新$\boldsymbol{w}$。

固定$\boldsymbol{M}$，式(8)可化简为：

$$\min_{\boldsymbol{w}} \boldsymbol{w}^{\mathrm{T}}\boldsymbol{P}\boldsymbol{w} - 2\boldsymbol{w}^{\mathrm{T}}\boldsymbol{q} \tag{19}$$

$$\mathrm{s.t.}\quad \sum_{i=1}^{4} w_i = 1, w_i \geqslant 0$$

其中：$\boldsymbol{q} \in \mathbf{R}^{4\times 1}$，$q_i = \mathrm{Tr}(\boldsymbol{M}^i\boldsymbol{M})$；$\boldsymbol{P} \in \mathbf{R}^{4\times 4}$，$p_i = \mathrm{Tr}(\boldsymbol{M}^i\boldsymbol{M}^j)$，由于$\sum_{i=1}^{4} w_i = 1$，这是一个线性约束的凸二次规划问题，可以用现有的优化工具求解。

2)固定$\boldsymbol{w}$，更新$\boldsymbol{M}$。

固定$\boldsymbol{w}$，式(1)可化简为：

$$\min_{\boldsymbol{M}} \|\boldsymbol{M} - \boldsymbol{C}\|_{\mathrm{F}}^2 \tag{20}$$

$$\mathrm{s.t.}\quad \boldsymbol{M} \geqslant 0, \boldsymbol{M} = \boldsymbol{M}^{\mathrm{T}}, \boldsymbol{M}\mathbf{1} = \mathbf{1}$$

其中，$\boldsymbol{C} \in \mathbf{R}^{n\times n}$，$\boldsymbol{C} = \sum_{i=1}^{d} w_i\boldsymbol{M}_d - \lambda\boldsymbol{D}$，$\boldsymbol{D}$是$\boldsymbol{F}$的欧氏距离矩阵。

根据式(16)可以得到：

$$\boldsymbol{M} = \boldsymbol{K} + \frac{n + \mathbf{1}^{\mathrm{T}}\boldsymbol{K}\mathbf{1}}{n^2}\mathbf{1}\cdot\mathbf{1}^{\mathrm{T}} - \frac{1}{n}\boldsymbol{K}\mathbf{1}\cdot\mathbf{1}^{\mathrm{T}} - \frac{1}{n}\mathbf{1}\cdot\mathbf{1}^{\mathrm{T}}\boldsymbol{K} \tag{21}$$

将$\boldsymbol{M}$的值作为输入代入式(20)中赋予$\boldsymbol{C}$，解易得：

$$\boldsymbol{M} = \max(0, \boldsymbol{C}) \tag{22}$$

3)固定$\boldsymbol{M}$，$\boldsymbol{w}$更新$\boldsymbol{F}$。

$$\min_{\boldsymbol{F}} \mathrm{Tr}(\boldsymbol{F}^{\mathrm{T}}\boldsymbol{L}\boldsymbol{F}) \tag{23}$$

$$\mathrm{s.t.}\quad \boldsymbol{F} \in \mathbf{R}^{n\times c}, \boldsymbol{F}^{\mathrm{T}}\boldsymbol{F} = \boldsymbol{I}$$

其中，$\boldsymbol{F}$为$\boldsymbol{L}$前$c$个最小的特征向量。

下面对目标函数求解过程进行总结。

求解式(1)、(3)、(5)的算法流程算法1所示。

算法1 加权优化。

输入 相似性矩阵信息$\left\{\boldsymbol{A}^i\right\}_{i=1}^{k}$；

输出 $\hat{\boldsymbol{S}}^d$。

初始化权重：$\boldsymbol{w}$；

重复

1)根据式(10)，更新$\hat{\boldsymbol{S}}^d$；

2)根据式(12)，更新$\boldsymbol{w}$；

直到$\hat{\boldsymbol{S}}^d$收敛。

求解式(2)、(4)、(6)、(7)的算法流程如算法2所示：

算法2 式(1)的优化算法。

输入 结构化矩阵：$\hat{\boldsymbol{S}}^d$；

输出 $\boldsymbol{M}^d$。

初始化自适应参数$\lambda$，初始化$\boldsymbol{F}$；

重复：

1)根据式(16)、(17)，迭代更新$\boldsymbol{M}^d$；

2)根据式(18)更新$\boldsymbol{F}$；

直到$\boldsymbol{M}^d$收敛。

求解式(8)的算法流程算法3所示。

算法3　式(8)的优化算法。

输入　$\{\boldsymbol{M}^d\}_{d=1}^4$

输出　$\boldsymbol{M}$。

初始化：$\boldsymbol{w}, \lambda, \boldsymbol{F}$；

重复：

1)根据式(19)，更新$\boldsymbol{w}$；

2)根据式(21)、(22)，迭代更新$\boldsymbol{M}$；

3)根据式(23)更新$\boldsymbol{F}$；

直到$\boldsymbol{M}$收敛。

## 3　实验与结果分析

### 3.1　数据集

本文使用以下8种不同类型的数据集进行聚类集成实验：1)CSTR(http://www.ncdc.ac.cn/portal/metadata/1a0e7fc8-8dc1-4c74-a6b2-e6a20d7b6ee4)；2)GLIOMA(https://sites.google.com/site/feipingnie/file/)；3)Prostate(https://cdas.cancer.gov/datasets/plco/20/)；4)ORL(http://www.uk.research.att.com/facedatabase.html)；5)YALE(http://cvc.yale.edu/projects/yalefaces/yalefaces.html)；6)Tr41(http://www.cs.umn.edu/~karypis/cluto/files /datasets.tar.gz)；7)Jaffe(http://www.kasrl.org/jaffe.html)；8)AR(http://www2.ece.ohio-state.edu/~aleix/ARdatabase.html)。使用不同类型的数据集可以更好地评估算法性能，数据集的详细信息如表1所示。

**表1　数据集详细信息**

Tab. 1　Detailed information of datasets

| 数据集 | 样本数 | 特征数 | 类数 |
|---|---|---|---|
| CSTR | 476 | 1 000 | 4 |
| GLIOMA | 50 | 4 434 | 4 |
| Prostate | 414 | 6 429 | 9 |
| ORL | 400 | 1 024 | 40 |
| YALE | 165 | 1 024 | 15 |
| Tr41 | 878 | 7 454 | 10 |
| Jaffe | 213 | 676 | 10 |
| AR | 840 | 768 | 120 |

### 3.2　对比方法

实验对比了以下9种算法：

1)K-Means(简写为KM)：表示$K$均值聚类的结果。聚类集成通常使用该算法作为基线。

2)CSPA(Cluster-based Similarity Partitioning Algorithm)[3]：表示了同一个簇种样本的关系，用于度量样本对之间的相似度。

3)HGPA(HyperGraph Partitioning Algorithm)[3]：一种基于超图划分邻域的方法，将超图的超边以及顶点所有的权重设为统一值。设置分区大小以避免出现大量碎片分区。

4)MCLA(Meta-CLustering Algorithm)[3]：该算法将聚类集成问题转化为簇一致性问题。

5)LWEA(Locally Weighted Evidence Accumulation)[23]：层次聚类方法，基于集成不确定估计和局部加权策略。

6)LWGP(Locally Weighted Graph Partitioning)[23]：一种基于局部加权策略的图划分算法；此外，通过熵的准则判断簇的可靠性。

7)RSEC(Robust Spectral Ensemble Clustering)[24]：一种具有鲁棒性的谱聚类方法。

8)DREC(Dense Representation Ensemble Clustering)[2]：该算法利用密集表示模型构造样本相似性矩阵。

9)SPEC(Self-Paced Clustering Ensemble)[25]：该方法从易到难进行学习，并将难度评估和集成学习融合在统一的框架中。

### 3.3　评价指标

本文实验采用聚类准确率(ACCuracy, ACC)和归一化互信息(Normalized Mutual Information, NMI)两种常见的聚类有效性外部评价指标评估算法性能。

ACC用于比较获得的标签和数据提供的真实标签，用$V_{\mathrm{ACC}}$表示，取值范围是[0,1]，值越大说明获得的标签准确性越高，将样本正确划分的效果越好。

$$V_{\mathrm{ACC}} = \frac{1}{n}\sum_{i=1}^{n}\delta\big(q_i, \mathrm{map}(p_i)\big) \tag{24}$$

其中：$p_i$是聚类后的标签；$q_i$是真实标签；$n$为样本总数。$\delta$表示指示函数，具体如下：

$$\delta(x, y) = \begin{cases} 1, & x = y \\ 0, & 其他 \end{cases} \tag{25}$$

NMI度量聚类结果的相似性程度，取值范围为[0,1]，值越大，说明变量之间的关系越密切，聚类结果越相近：

$$NMI(\boldsymbol{A}, \boldsymbol{B}) = \frac{I(\boldsymbol{A}, \boldsymbol{B})}{\big(H(\boldsymbol{A}) + H(\boldsymbol{B})\big)/2} \tag{26}$$

其中：$H(\boldsymbol{A})$、$H(\boldsymbol{B})$是$\boldsymbol{A}$、$\boldsymbol{B}$的熵；$I(\boldsymbol{A}, \boldsymbol{B})$是互信息，表示一个变量包含另一个变量的信息量；$\boldsymbol{A}$是真实数据的标签集合，$\boldsymbol{B}$是聚类算法划分的类集合。互信息$I(\boldsymbol{A}, \boldsymbol{B})$表示为：

$$I(\boldsymbol{A}, \boldsymbol{B}) = \sum_{a_i \in \boldsymbol{A}, b_i \in \boldsymbol{B}} p(a_i, b_i)\,\mathrm{lb}\frac{p(a_i, b_i)}{p(a_i)p(b_i)} \tag{27}$$

其中：$p(a_i)$为从数据集中任意选定一个样本点属于$\boldsymbol{A}$类的概率；$p(a_i, b_i)$为任选的数据点同时属于$\boldsymbol{A}$类和$\boldsymbol{B}$类的概率。

### 3.4　实验结果与分析

本文将通过实验验证高阶信息以及高阶信息融合的有效性。不同算法在8个数据集上的结果对比如表2~4所示，其中：最优结果加粗表示；次优结果用下划线表示；括号中的数据为方差。

表2为不同算法的ACC结果，可以看出：HCLCE算法在一定程度上提高了聚类集成的聚类效果，在不同数据集上的实验结果大部分高于对比算法；并且HCLCE算法相比其他对比算法，具有较小的方差，说明HCLCE算法的稳定性更好。对比鲁棒性方法RSEC，HCLCE算法具有更好的表现。

表3为不同算法的NMI结果对比。从表2~3可以看出，HCLCE算法的ACC和NMI在所有数据集上的均值均好于对比算法。与次优的LWEA相比，ACC平均提升了7.22%，NMI平均提升了9.19%。

HCLCE算法融合多种高阶信息，在多数情况下好于仅使用一种信息的聚类结果。使用不同高阶信息矩阵$\boldsymbol{A}$作为输入，进行加权结构化后得到新的关联矩阵$\boldsymbol{M}$。表4为不同的$\boldsymbol{M}$在融合前的聚类效果和融合后整体的聚类效果。其中

$\boldsymbol{A}^i$的定义已在前面介绍，不同的集合代表着不同的高阶信息计算方式，集合从大小到所表示信息具有很大差异性。$\boldsymbol{M}^1$是加权结构化后的一阶信息关联矩阵，以$\boldsymbol{M}^1$为基础进行聚类，效果比对比方法有一定提升，说明对不同输入加权起到了让质量好的输入权重大、质量差的输入权重小的作用，从而提高聚类结果。并且结构化和秩约束使样本对关系表达得更加清楚，簇的结构更加清晰。融合的过程再次对不同阶信息分配权重，使各种信息再次组合。

**表2 ACC实验结果对比**

Tab. 2 Comparison of ACC experimental results

| 数据集 | KM | CSPA | HGPA | MCLA | LWEA | LWGP | RSEC | DREC | SPCE | HCLCE |
|---|---|---|---|---|---|---|---|---|---|---|
| AR | 0.3301 (0.087) | 0.355 (0.011) | 0.3807 (0.012) | 0.3337 (0.115) | 0.3898 (0.013) | 0.3645 (0.013) | 0.2938 (0.006) | 0.4023 (0.007) | 0.3499 (0.007) | **0.4136** (0.006) |
| CSTR | 0.7331 (0.087) | 0.6804 (0.038) | 0.2897 (0.032) | 0.7966 (0.029) | 0.8019 (0.004) | 0.8432 (0.057) | 0.8589 (0.074) | 0.8293 (0.071) | 0.8046 (0.008) | **0.9019** (0.009) |
| GLIOMA | 0.4292 (0.037) | 0.4220 (0.033) | 0.4360 (0.031) | 0.4080 (0.014) | 0.4320 (0.021) | 0.4100 (0.030) | 0.4000 (0.041) | 0.434 (0.010) | 0.4340 (0.030) | **0.4420** (0.014) |
| Prostate | 0.7402 (0.068) | 0.6517 (0.012) | 0.5618 (0.012) | 0.7034 (0.014) | 0.6978 (0.004) | 0.6989 (0.007) | 0.6931 (0.085) | 0.5506 (0.069) | 0.6978 (0.068) | **0.8076** (0.076) |
| Jaffe | 0.7603 (0.087) | 0.9286 (0.040) | 0.8939 (0.048) | 0.9333 (0.043) | 0.9338 (0.046) | 0.8282 (0.086) | 0.7906 (0.065) | 0.9277 (0.055) | 0.8803 (0.029) | **0.9606** (0.013) |
| ORL | 0.4859 (0.032) | 0.5720 (0.025) | 0.5768 (0.021) | 0.5873 (0.012) | 0.5735 (0.021) | 0.5328 (0.031) | 0.375 (0.019) | **0.6090** (0.024) | 0.5310 (0.066) | 0.5930 (0.019) |
| YALE | 0.3678 (0.034) | 0.3915 (0.024) | 0.4006 (0.022) | 0.4067 (0.021) | 0.4048 (0.024) | 0.4097 (0.027) | 0.2776 (0.037) | 0.4346 (0.026) | 0.3655 (0.016) | **0.4436** (0.020) |
| Tr41 | 0.5709 (0.072) | 0.5093 (0.029) | 0.4687 (0.033) | 0.5726 (0.046) | 0.6872 (0.053) | 0.6535 (0.037) | 0.6309 (0.054) | 0.6500 (0.035) | 0.6695 (0.087) | **0.7136** (0.045) |
| 平均 | 0.5522 (0.066) | 0.5638 (0.028) | 0.5010 (0.032) | 0.5927 (0.047) | 0.6151 (0.034) | 0.5926 (0.047) | 0.5399 (0.068) | 0.6046 (0.046) | 0.5915 (0.038) | **0.6595** (0.031) |

**表3 NMI实验结果对比**

Tab. 3 Comparison of NMI experimental result

| 数据集 | KM | CSPA | HGPA | MCLA | LWEA | LWGP | RSEC | DREC | SPCE | HCLCE |
|---|---|---|---|---|---|---|---|---|---|---|
| AR | 0.6390 (0.064) | 0.7015 (0.004) | 0.7039 (0.006) | 0.6878 (0.005) | 0.6748 (0.007) | 0.6825 (0.009) | 0.5828 (0.015) | 0.6911 (0.007) | **0.7279** (0.002) | 0.7046 (0.005) |
| CSTR | 0.6390 (0.064) | 0.5037 (0.041) | 0.0150 (0.016) | 0.6734 (0.019) | 0.6902 (0.008) | 0.7183 (0.043) | 0.7526 (0.044) | 0.7100 (0.071) | 0.6703 (0.018) | **0.7718** (0.021) |
| GLIOMA | 0.1673 (0.040) | 0.1760 (0.037) | 0.1651 (0.023) | 0.1508 (0.030) | 0.1605 (0.022) | 0.1469 (0.030) | 0.1061 (0.036) | 0.1705 (0.009) | 0.1550 (0.028) | **0.1820** (0.020) |
| Prostate | 0.1637 (0.091) | 0.0810 (0.013) | 0.1280 (0.005) | 0.1124 (0.013) | 0.1073 (0.003) | 0.1073 (0.003) | 0.1183 (0.075) | 0.0803 (0.084) | 0.1073 (0.030) | **0.2577** (0.104) |
| Jaffe | 0.4718 (0.087) | 0.9105 (0.033) | 0.8834 (0.041) | 0.9234 (0.028) | 0.9225 (0.029) | 0.8775 (0.040) | 0.8408 (0.059) | 0.9318 (0.055) | 0.8738 (0.022) | **0.9473** (0.014) |
| ORL | 0.6898 (0.020) | 0.7499 (0.012) | 0.7616 (0.008) | 0.7534 (0.006) | 0.7616 (0.009) | 0.7270 (0.016) | 0.5860 (0.016) | **0.7741** (0.024) | 0.7663 (0.005) | 0.7597 (0.006) |
| YALE | 0.4206 (0.031) | 0.4432 (0.014) | 0.4475 (0.020) | 0.4482 (0.014) | 0.4381 (0.024) | 0.4522 (0.025) | 0.2996 (0.045) | 0.4866 (0.045) | 0.4570 (0.046) | **0.5027** (0.011) |
| Tr41 | 0.5896 (0.053) | 0.5874 (0.015) | 0.4805 (0.037) | 0.6184 (0.031) | 0.6773 (0.030) | 0.6620 (0.023) | 0.6456 (0.037) | 0.6639 (0.035) | 0.6128 (0.096) | **0.7136** (0.025) |
| 平均 | 0.4726 (0.059) | 0.5191 (0.025) | 0.4481 (0.029) | 0.5459 (0.087) | 0.5540 (0.024) | 0.5467 (0.030) | 0.4914 (0.073) | 0.5635 (0.069) | 0.5463 (0.058) | **0.6049** (0.034) |

**表4 信息融合前后不同阶的ACC对比**

Tab. 4 ACC Comparison at different leves before and after information fusion

| 数据集 | 不同阶的ACC | | | | |
|---|---|---|---|---|---|
| | $\boldsymbol{M}^1$ | $\boldsymbol{M}^2$ | $\boldsymbol{M}^3$ | $\boldsymbol{M}^4$ | $\boldsymbol{M}$ |
| AR | 0.4127(0.009) | 0.4046(0.008) | 0.4152(0.005) | 0.2623(0.094) | **0.4136**(0.006) |
| CSTR | 0.9002(0.009) | 0.8783(0.042) | 0.8994(0.011) | 0.4752(0.111) | **0.9019**(0.009) |
| GLIOMA | 0.4440(0.015) | 0.4340(0.017) | 0.4380(0.015) | 0.4080(0.055) | **0.4420**(0.014) |
| Prostate | 0.7607(0.084) | 0.7753(0.084) | 0.7753(0.083) | 0.7213(0.105) | **0.8076**(0.076) |
| Jaffe | 0.9469(0.041) | 0.9333(0.052) | 0.9455(0.040) | 0.4545(0.082) | **0.9606**(0.013) |
| ORL | 0.5860(0.007) | 0.5815(0.032) | 0.5625(0.012) | 0.2240(0.005) | **0.5930**(0.019) |
| YALE | 0.4194(0.014) | 0.4145(0.020) | 0.4158(0.015) | 0.2473(0.062) | **0.4436**(0.020) |
| Tr41 | 0.6779(0.038) | 0.6426(0.003) | **0.7204**(0.043) | 0.3605(0.026) | 0.7136(0.045) |
| 平均 | 0.6698(0.035) | 0.6581(0.040) | 0.6794(0.037) | 0.4420(0.053) | **0.6843**(0.031) |

每种高阶信息从不同的角度表示样本对相似性,因此有不同的特点。以 CSTR 数据集为例,不同阶信息表示的关联矩阵经过加权结构化后的直观展示如图 5 所示:颜色越深,样本相似性越小;颜色越浅,样本相似性越大。

从图 5 可以看出:1)使用原始输入的聚类结果得到的结构化相似性矩阵 $\boldsymbol{M}^1$ 中,大量样本对之间相似性处于 0.5~0.6,很难判断两个样本是否属于同一类。2)对于矩阵叉乘 $\boldsymbol{M}^2$,得到的相似性矩阵区分度不高,大量样本对同时具有高相似度,这种信息过于冗余,基聚类输入两两之间在簇上的产生交集的概率很大,特别是在基聚类器之间差异性不大的情况下,同样不具有区分性。3)矩阵样本对之间的一致性介于原始输入和簇级一致性之间,说明基聚类器两两之间在样本对一致性判断上不能统一,有的簇中样本对一致性较大,有些簇中样本对一致性趋于二分,不容易判断。4)单独使用所有关联矩阵点积连乘运算获得的 $m$ 阶信息的聚类效果明显下降。这是因为 $m$ 阶信息虽然可靠但非常稀疏,只保留了所有输入达成共识的样本对,没有保留一致性较大的样本对,关联不好的簇作为输入会影响整体聚类的效果。

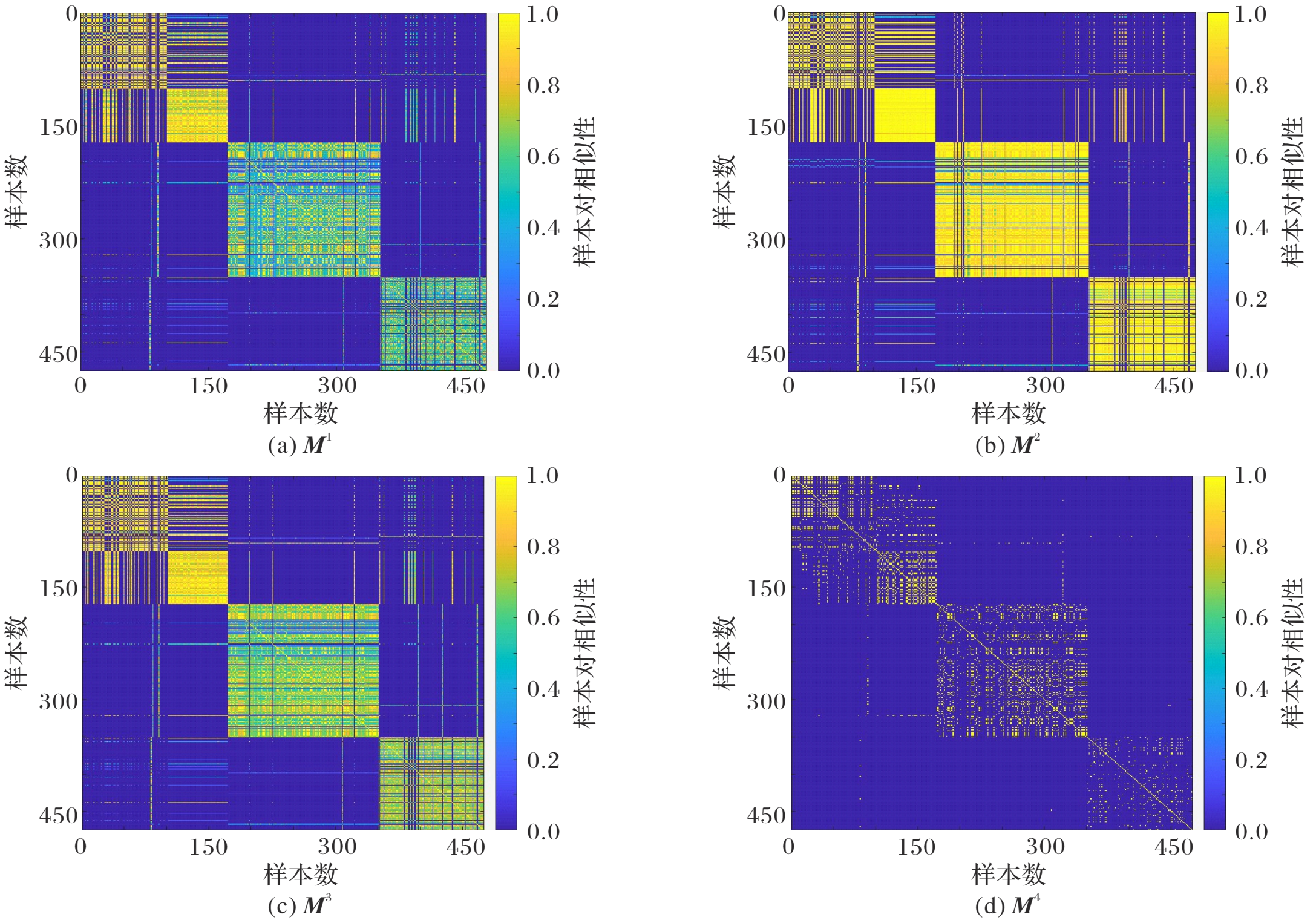

图 5　不同阶数信息的结构化关联矩阵

Fig. 5　Structured correlation matrices of different order information

HCLCE 算法对融合的每种信息赋予权重:与目标差异越大的信息获得的权重越小,减轻了不好的基聚类器带来的影响;质量高的基聚类器占据主导地位,提高了最后的聚类结果。经过高阶信息融合后得到的关联矩阵簇结构更清晰,去除了很多冗余样本对信息,更加满足关联矩阵的性质。

## 4　结语

本文提出了一种新的数据高阶信息挖掘方法,利用高阶一致性共识的信息,从不同角度刻画样本之间的联系,验证了不同层面的共识信息的差异性。HCLCE 算法通过加权减少信息之间的质量差异性带来的影响,引入对关联矩阵双随机约束和秩约束,使得最终融合的关联矩阵更加符合其内在特性。通过对多种高阶信息的融合,得到了比聚类集成算法和单独使用一种信息更好的聚类结果。实验结果表明,差异性大的输入对于聚类结果的提升具有帮助。其次,通过实验验证了每一种信息的特点和有效性,以及融合算法要好于单独使用某一种信息。此外观察到 $m$ 阶信息虽代表了可信度最高的一致性样本对信息,但是在融合过程中没有起到明显的提升效果或者是约束样本对一致性的监督作用。在后续工作中,应探索在聚类过程中如何充分利用可靠信息,从可靠信息中发掘样本潜在的一致性信息,从而更大程度地减少低质量信息对聚类结果产生的负面影响。

## 参考文献 (References)

This work is partially supported by National Natural Science Foundation of China (61976129).

**GAN Jianwen**, born in 1996, M. S. candidate. His research interests include clustering ensemble, data mining.

**CHEN Yan**, born in 1994, Ph. D. candidate. Her research interests include multi-core clustering, deep clustering.

**ZHOU Peng**, born in 1989, Ph. D., associate professor. His research interests include clustering ensemble, data mining.

**DU Liang**, born in 1985, Ph. D., associate professor. His research interests include machine learning, big data analysis.